# Problem of robotic precision cutting of the geometrically complex shape from an irregular honeycomb grid

M.V. Kubrikov[1], M.V. Saramud[1,2][0000-0003-0344-9842] and M.V Karaseva[1,2]

[1] Reshetnev Siberian State University of Science and Technology, Krasnoyarsky Rabochy Av. 31, Krasnoyarsk, 660037, Russian Federation
[2] Siberian Federal University, 79 Svobodny avenue, Krasnoyarsk, 660041, Russian Federation
msaramud@gmail.com

**Abstract.** The article considers solving the problem of precision cutting of honeycomb blocks. The urgency of using arbitrary shapes application cutting from honeycomb blocks made of modern composite materials is substantiated. The problem is to obtain a cut of the given shape from honeycomb blocks. The complexity of this problem is in the irregular pattern of honeycomb blocks and the presence of double edges, which forces an operator to scan each block before cutting. It is necessary to take into account such restrictions as the place and angle of the cut and size of the knife, its angle when cutting and the geometry of cells. For this problem solving, a robotic complex has been developed. It includes a device for scanning the geometry of a honeycomb block, software for cutting automation and a cutting device itself. The software takes into account all restrictions on the choice of the location and angle of the operating mechanism. It helps to obtain the highest quality cut and a cut shape with the best strength characteristics. An actuating device has been developed and implemented for both scanning and cutting of honeycomb blocks directly. The necessary tests were carried out on real aluminum honeycomb blocks. Some technical solutions are used in the cutting device to improve the quality of cutting honeycomb blocks. The tests have shown the effectiveness of the proposed complex. Robotic planar cutting made it possible to obtain precise cutting with a high degree of repeatability.

## 1 Introduction

The development of industry, especially aviation, rocket and space technologies, is closely related to the development of new design and technological solutions based on modern materials, including high-strength fibrous polymer composite materials [1]. A great attention is paid to the efficiency of modern spacecrafts; it leads to the necessity to search for new technological solutions [2]. The application of three-layer structures is the main and most important field of research.

Three-layer structures are two load-bearing faces and lightweight core materials located between them. This lightweight core material is mainly honeycomb [3]. Load-bearing faces perceive longitudinal loads (tension, compression, and shear) in their plane and transverse bending moments [4]. The lightweight core material absorbs the shear forces at bending of the three-layer structure and ensures the joint work and stability of the load-bearing faces. The honeycomb core is designed to produce lightweight, rigid and heat-insulating panels [5]. Due to its characteristics, this type of structure has found active use in spacecraft designs, since it provides a significant reduction in the mass of their structures, and, as a consequence, an increase in the mass efficiency of spacecraft's in general. In this regard, space platforms of various classes have been developed and they are actively used on the basis of structures made of honeycomb panels [6-7].

The cost of the main structural lightweight core materials is very high, and the production of curved panels or panels of variable thickness on their basis is a complex technological problem. The application of multilayer structures is increasing. Scientific and practical problems associated with them are of a great importance [8].

The article is devoted to solving the urgent problem of producing structures using honeycomb blocks, namely, to the solution of automated precision cutting of honeycomb blocks from various materials.

## 2    Problem of cutting honeycomb blocks

The application of a honeycomb block of aluminum, aramid or low-thickness carbon fiber as the main material does not allow using classical methods of material processing, such as milling. A shear line and nearby cells are compressed when milling is a thin jumper of a honeycomb block. The main problem when cutting such material is passing the cutting tool along a path that does not take into account the honeycomb structure of the block. Passing the cutting tool through the points connection of the cells often leads to the cells deformation. Also it creates an unreinforced cell wall. A wall without reinforcement with other cells is not able to withstand the applied loads. As a result, an edge of the sandwich panel will have less strength in comparison with the main surface.

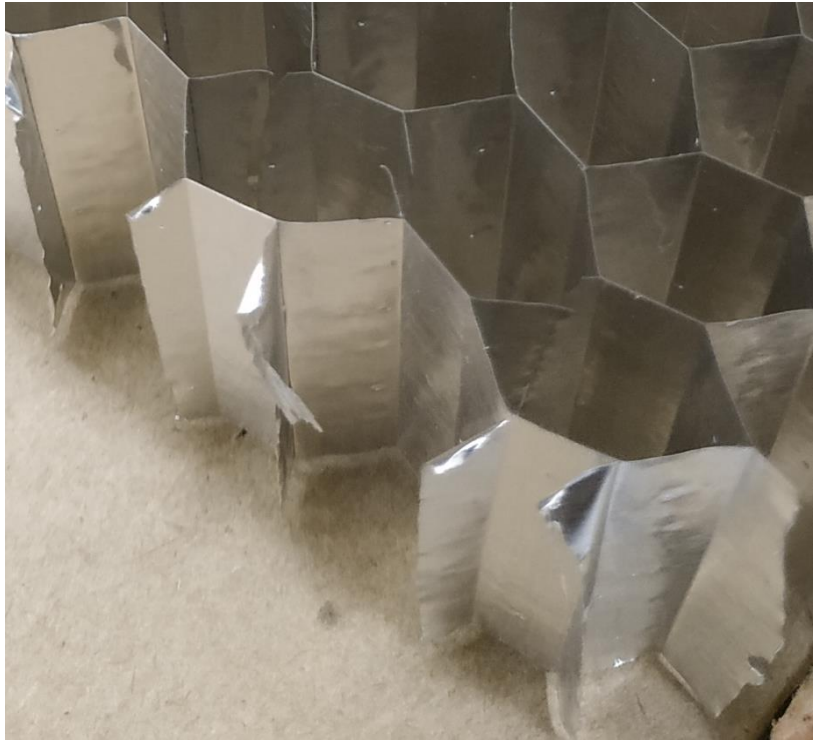

**Fig. 1.** Aluminum honeycomb block

The technology for obtaining a high-quality cut of using a special knife is investigated. It makes cutting a honeycomb block of complex geometry, along a trajectory with a minimum radius of curvature equal to the cell size possible. Figure 1 presents cutting a wall far away from the place of gluing. This cutting leads to its strong deformation. This effect can be reduced by moving the cut closer to the junction. Figure 1 shows a manufacture cut along the outer contour of the block and the cut, presented by authors of the article, i.e., it is located one cell from an edge.

We are dealing with an aluminum honeycomb block with a height of 40 mm and a cell edge length of 10 mm. A photo of a honeycomb block is given in Figure 1.

The method of cutting with a special knife was chosen as the most suitable for the problem of cutting honeycomb blocks solving.

Four main problems when cutting edges of honeycomb blocks arise. They are a choice of the correct angle of the knife blade relative to the cut edge, a choice of the cutting place on the edge relative to the nodal points of intersection of the edges, a choice of single edges and the location of the knife in space to exclude unwanted contact with adjacent edges. The cutting edges process is shown schematically in Figure 2; glued edges with double thickness are shown in blue. The honeycomb structure is obtained by gluing and further extension of aluminum strips. Thus, edges of double thickness are formed.

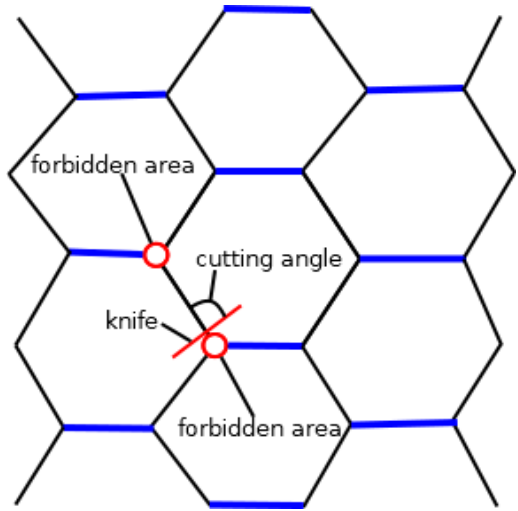

**Fig. 2.** Representation of the edge cutting in the honeycomb block.

It is required to automate the determination of points and cutting angles to eliminate the described problems. It is necessary to obtain a "map" of a real honeycomb block to solve this problem. For this, a honeycomb block must be fixed on the surface where the cutting will be performed. A scanning device should be installed and synchronized with the software. And then a honeycomb block must be scanned.

The algorithm is required; it will optimally locate a cut shape on the surface of the honeycomb block, taking into account the described requirements. The algorithm selects the optimal location of the knife at each point of intersection of the shape and the edges of the honeycomb block. As a result, one gets an array of cutting points with an indication of the angle of the knife. All these points meet our requirements, i.e., a cut is no closer than 0.5 mm from the intersection points of the edges, a knife is at the optimal angle to the cut edge and it does not touch adjacent edges. If it is possible, a cut passes along single edges, as close as possible to the nodal points. This cutting will help to get the most reliable structure.

After that, the software converts the resulting data array into a G-code that can be understood by the equipment.

## 3       Honeycomb block scanning

A honeycomb block located on the working surface of the robotic complex is used as an initial object. Due to the high flexibility of the honeycomb structure of the block, resulting from the small thickness of the partitions, the cells have different geometries. The use of technical vision is a necessary measure to obtain cutting only along the edges of the honeycomb, without destroying the nodal joints.

A scanning process begins with the sequential passage of the actuator with the installed video camera over the surface of the honeycomb block. The resulting video sequence is formed into a panoramic image of the entire surface of the honeycomb block.

Nodal points (intersection of several edges) and honeycomb edges are formed from a panoramic image with the help of technical vision. Technical vision helps to bind the coordinates of the nodal points to the coordinates of the working surface of the robotic complex. Thereby it is possible to obtain a digital copy of a real honeycomb block. Also, double and single edges are marked on the map. The process is simplified by their arrangement, i.e., double edges are formed as a result of gluing aluminum strips and they are located on one straight line.

The open CV library is used to select the required objects [9]; it is used in the languages c ++, python, java. A neural network is used to detect the required objects. The TensorFlow [10] and Keras [11] libraries are applied to train it. The python programming language is applied for its implementation.

As a result of this operation, it is possible to get a "map" of our honeycomb block, with the real dimensions and positions of all the node intersection points of the faces, marked with double faces.

## 4 Software implementation of the proposed approach

So, there exists the scanned geometry of a honeycomb block in real dimensional units, node points of intersection of faces, information about each face, i.e., single or glued. The form that needs to be cut is loaded into the program. First, a check for sufficient size is carried out. If a form cannot be positioned on the block even not taking into account the restrictions, the program generates an error, informing a user about it. If the size of the block is sufficient for the selected shape, selection of the optimal location of the shape on it begins. Then, a shape is aligned along the longest straight line, if any, and it is located parallel to the block so that this straight line passes along the single edges at the optimal distance from the intersection points. The original location of the form is in the lower left corner for the most rational application of the material. The program also specifies whether an existing block edge can be applied. It will greatly simplify the process if the existing block edge is cut cleanly and smoothly, and one of the sides of the shape is straight. In this case, the edge of the form is aligned with the edge of the block and further movement is carried out in the same plane.

Further, all points of intersection of the shape and honeycomb structure are checked. Each point is first checked for coincidence with the unwanted area, an area with a radius of 0.5 mm around the anchor points of intersection of edges or double edges if there are such intersections. The point is placed as a problematic one and a shape is moved. If there are several problem points, the search for suitable places for cutting within a radius of 5 mm from these points is carried out.

When the location of the shape is found without problem points, the further verification is performed. At each point, the contour of the knife is located at the optimal angle to the cut face and it is checked whether it touches adjacent edges. If there exist intersections, a knife angle changes within acceptable values to ensure a correct cutting. If changing the angle eliminates the intersection problem, it is stored by the program, if not, a point is marked as problematic and the shape changes its location again.

These iterations continue until a location is found where there are no problem points. As a result, we get an array of cutting points, i.e., their coordinates and knife angles for each of them. The software generates the G-code required to control a operating mechanism.

The proposed algorithm is implemented in the form of a software environment for the automation of cutting honeycomb blocks. The interface of the software system is shown in Figure 3.

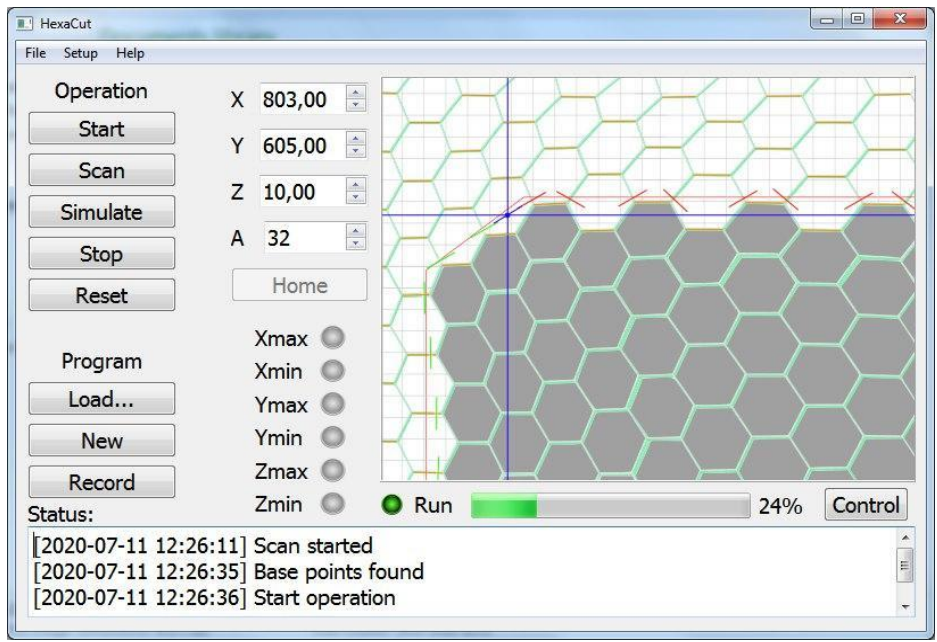

**Fig. 3.** Software system interface

Having fixed a honeycomb block on the cutting surface, it moves a operating mechanism to the zero point (Home). It helps to apply operating mechanism coordinates for binding the geometry of the honeycomb cell structure.

After that, use a button Load to upload the required pattern. The pattern file contains an array of vectors. The program searches for closed contours and visualizes the beginning of the cutting and its direction.

By clicking on a button Scan, a cell scanning process starts. It is necessary to take into account the parameters of the honeycomb block in order to control the height of the machine vision camera. Linear scanning produces a series of images to obtain an overall picture of the entire honeycomb blocks. The complete image is sent for honeycomb geometry recognition and vectorized image acquisition.

The recognized vectors are transferred back to the software. The automated algorithm seeks to locate optimally the pattern on an irregular honeycomb structure, while the pattern should not go beyond the surface of the honeycomb block. The next step in the algorithm realization is to find the cutting points, while several conditions are met. The main is that a cutting point must be outside the closed contour and at least 0.5 mm from the nodal point of the cell. Also it is necessary to take into account all the other restrictions described above. At this stage, the selected points and the location of the knife are visualized. Clicking a button Simulate the G-code is generated and a simulation of the operating mechanism is displayed in the graphics window. A button Start is activated upon completion of the simulation. So, one can start the fulfillment of the operating mechanism for the actual operation.

## 5    Operating mechanism for cutting

An actuating unit, an operating mechanism, has been developed for cutting edges of honeycomb blocks and machine vision for the practical application of the described approach. A photograph of such a operating mechanism is given in Figure 4.

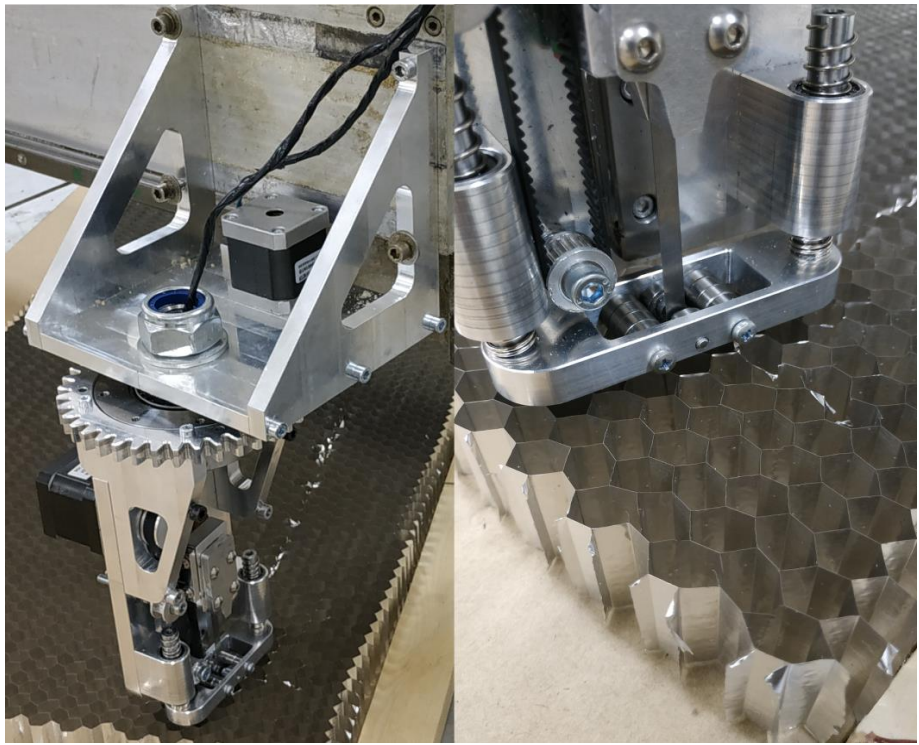

**Fig. 4.** Operating mechanism for cutting

The operating mechanism consists of several autonomous systems on a rotary mechanism. A rotary mechanism is made on radial axial bearings with a central hollow rod. It is driven by a stepper motor with the ability to rotate with an accuracy of 0.1 degree. The main mounting plate is attached to the rotate mechanism. A machine vision camera is fixed on its one side. A mechanism for vertical movement of the knife is on the other side. The knife speed reaches 1500 mm/sec. The cutting knife is made of tool steel; its thickness is only 0.3 mm. The knife's width is 5 mm. A photo of the knife profile is given in Figure 5.

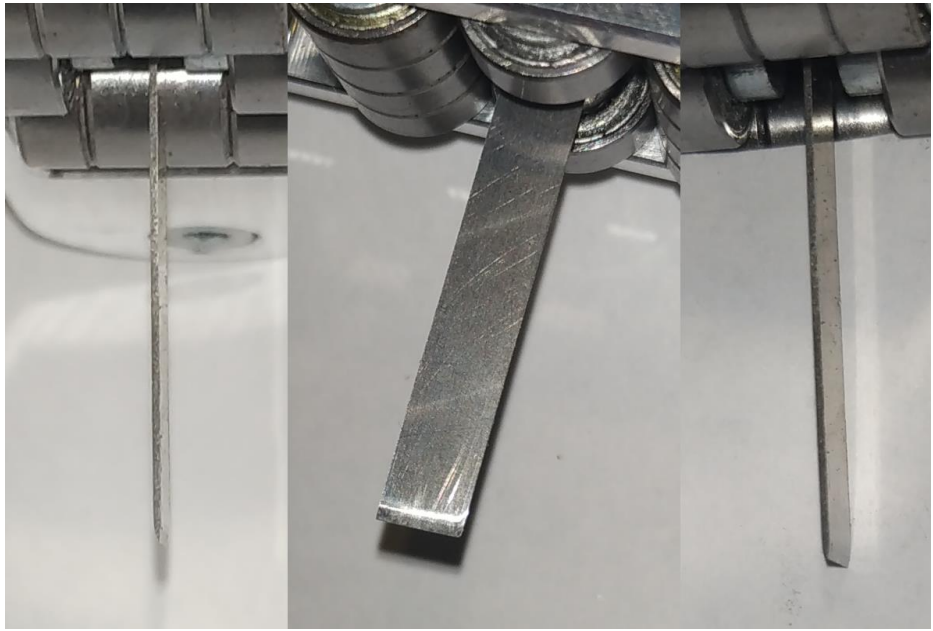

**Fig. 5.** Cutting knife

A polyurethane backing is installed on the working surface. Its use allows reducing knife wear and increasing the influence of knife positioning errors.

The operating mechanism has a spring-loaded platform, ensuring that the honeycomb block is pressed to the surface. Due to the high flexibility of the material, pressing is necessary to ensure that the honeycomb block remains in position at the time the knife is removed.

## 6  Result of the system operation

The developed robotic complex was tested on a real aluminum honeycomb block. Seven different types of shapes were cut, including sections of different curvature and sections with a change in the generatrix at right angles. There are also holes for em-bedded fasteners with a diameter of 40 mm. The cell edge of the honeycomb block is 10 mm and its thickness is 100 μm.

There were five runs for each form. The accuracy of determining the geometry of the cells was 0.1 mm as a result of 35 experiments. The positioning accuracy of the knife relative to the cellular structure of the block was no more than 0.1 mm. Meanwhile, the scanning speed was 1000 mm2/s, and the cutting speed was 5 mm/s. The average area of the object is 0.17 m2.

The maximum indentation while maintaining the integrity of the cellular structure from the boundary of the object is 5 mm. This spread is easily compensated for in the finished product due to the high elasticity of the honeycomb block.

Thus, the obtained edge of the honeycomb block as a part of the finished sandwich panel has stable transverse strength. It has a positive effect when several panels are joined into a finished product.

## 9     Conclusion

The investigation in the field of the process of spatial precision cutting of honeycomb blocks made of composite materials, guarantees the development of high-quality lightweight structures in the aerospace and related industries.

The technical problem posed for robotic precision cutting of geometrically complex shapes from an irregular honeycomb grid was solved. As a result, a cutting algorithm was obtained. It provides a final shape with the best strength characteristics. The resulting algorithm is implemented in the form of a software and hardware complex. This complex includes a vision module, cutting automation, generation of a control code for the operating mechanism with an operating mechanism and the operating mechanism itself. It allows both scanning and cutting of honeycomb blocks.

A series of experiments made it possible to find the optimal solution for the development a special knife.

The robotic spatial cutting made it possible to obtain precise cutting with a high degree of repeatability.

**Acknowledgments.** This work was supported by the Ministry of Science and Higher Education of the Russian Federation (State Contract No. FEFE-2020-0017)

## References

1. Bayraktar, Emin: Section 12 Composites Materials and Technologies. 10.1016/B978-0-12-803581-8.04108-4 (2016).
2. Crupi, V., Epasto, G., Guglielmino, E.: Comparison of aluminium sandwiches for lightweight ship structures: honeycomb vs. foam, Marine Structures. vol. 30, pp. 74–96. (2013).

3. Meifeng He, Wenbin Hu: A study on composite honeycomb sandwich panel structure. Materials and Design, vol. 29, pp. 709–713. (2008).
4. Crupi, Vincenzo & Epasto, Gabriella & Guglielmino, E.: Collapse Modes in Aluminium Honeycomb Sandwich Panels Under Bending and Impact Loading. International Journal of Impact Engineering, vol. 43, pp. 6-15, 10.1016/j.ijimpeng.2011.12.002 (2012)
5. Hu, Zhong & Thiyagarajan, Kaushik Venugopalan & Bhusal, Amrit & Letcher, Todd & Fan, Qi & Liu, Qiang & Salem, David: Design of ultra-lightweight and high-strength cellular structural composites inspired by biomimetics. Composites Part B Engineering. 10.1016/j.compositesb.2017.03.033 (2017).
6. Boudjemai, A., Bouanane, M.H., Mankour, Amri, R., Salem, H., Chouchaoui, B.: MDA of Hexagonal Honeycomb Plates used for Space Applications. World Academy Of Science, Engineering And Technology Issue 66 June 2012, (2012).
7. Boudjemai, Abdelmadjid & Amri, R. & Mankour, Abdeldjelil & Salem, H. & Bouanane, M.H. & Boutchicha, Djilali: Modal analysis and testing of hexagonal honeycomb plates used for satellite structural design. Materials & Design, vol. 35, pp. 266–275, 10.1016/j.matdes.2011.09.012 (2012).
8. Regassa, Yohannes & Lemu, Hirpa & Sirabizuh, Belete: Trends of using polymer composite materials in additive manufacturing. IOP Conference Series: Materials Science and Engineering. 659. 012021. 10.1088/1757-899X/659/1/012021 (2019).
9. OpenCV Homepage, https://opencv.org, last accessed 2020/07/30.
10. TensorFlow Homepage, https://www.tensorflow.org, last accessed 2020/07/30.
11. Keras Homepage, https://keras.io, last accessed 2020/07/30.